\documentclass[lettersize,journal]{IEEEtran}
\usepackage{amsmath,amsfonts}
\usepackage{algorithmic}
\usepackage{algorithm}
\usepackage{array}
\usepackage[caption=false,font=normalsize,labelfont=sf,textfont=sf]{subfig}
\usepackage{textcomp}
\usepackage{stfloats}
\usepackage{url}
\usepackage{verbatim}
\usepackage{graphicx}
\usepackage{cite}
\usepackage[utf8]{inputenc}
\usepackage{graphicx}
\usepackage{amsmath}
\usepackage{amsthm}
\usepackage{amssymb}
\usepackage{mathrsfs}
\usepackage{booktabs}
\usepackage{color}
\usepackage{multirow}
\usepackage[switch]{lineno}
\usepackage{colortbl}
\usepackage{bbding}

\def\BibTeX{{\rm B\kern-.05em{\sc i\kern-.025em b}\kern-.08em
    T\kern-.1667em\lower.7ex\hbox{E}\kern-.125emX}}
\usepackage{balance}

\usepackage{colortbl} 
\usepackage{xcolor}   

\begin{document}

\title{C${^2}$RL: Content and Context Representation Learning for Gloss-free Sign Language Translation and Retrieval}

\author{Zhigang Chen*, Benjia Zhou*, Yiqing Huang, Jun Wan$\dagger$,~\IEEEmembership{Senior Member,~IEEE}, \\ Yibo Hu, Hailin Shi,~\IEEEmembership{Member,~IEEE}, Yanyan Liang,~\IEEEmembership{Member,~IEEE},  Zhen Lei,~\IEEEmembership{Fellow,~IEEE}, Du Zhang
\thanks{* Contributed equally to this work.}
\thanks{$\dagger$ Corresponding author.}
\thanks{Z. Chen, J. Wan and Z. Lei are with the State Key Laboratory of Multimodal Artificial Intelligence Systems (MAIS), Institute of Automation, Chinese Academy of Sciences (CASIA), Beijing 100190, China, also with the School of Artificial Intelligence, University of Chinese Academy of Sciences (UCAS), Beijing 100049, China. J. Wan is also with Macau University of Science and Technology, Macau 999078, China.
}
\thanks{B. Zhou, Y. Huang, Y. Liang, and D. Zhang are with Macau University of Science and Technology, Macau 999078, China.}
\thanks{Y. Hu and H. Shi are with NIO, Digital Cockpit and Software Development,
Beijing, China.}
\thanks{Manuscript received xx xx, 2024; revised xx xx, 2024.}
}

\markboth{Journal of \LaTeX\ Class Files,~Vol.~14, No.~8, August~2021}%
{Shell \MakeLowercase{\textit{et al.}}: A Sample Article Using IEEEtran.cls for IEEE Journals}


\maketitle

\def\ie{{\em i.e.}}
\def\eg{{\em e.g.}}
\def\vs{\emph{vs.}}
\def\etal{{\em et al.}}
\def\etc{{\em etc.}}
\def\X{\XSolidBrush}
\def\C{\CheckmarkBold}

\begin{abstract}

    Sign Language Representation Learning (SLRL) is crucial for a range of sign language-related downstream tasks such as Sign Language Translation (SLT) and Sign Language Retrieval (SLRet). Recently, many gloss-based and gloss-free SLRL methods have been proposed, showing promising performance. Among them, the gloss-free approach shows promise for strong scalability without relying on gloss annotations. However, it currently faces suboptimal solutions due to challenges in encoding the intricate, context-sensitive characteristics of sign language videos, mainly struggling to discern essential sign features using a non-monotonic video-text alignment strategy. Therefore, we introduce an innovative pretraining paradigm for gloss-free SLRL, called C${^2}$RL, in this paper. Specifically, rather than merely incorporating a non-monotonic semantic alignment of video and text to learn language-oriented sign features, we emphasize two pivotal aspects of SLRL: Implicit Content Learning (ICL) and Explicit Context Learning (ECL). ICL delves into the content of communication, capturing the nuances, emphasis, timing, and rhythm of the signs. In contrast, ECL focuses on understanding the contextual meaning of signs and converting them into equivalent sentences. Despite its simplicity, extensive experiments confirm that the joint optimization of ICL and ECL results in robust sign language representation and significant performance gains in gloss-free SLT and SLRet tasks. Notably, C${^2}$RL improves the BLEU-4 score by +5.3 on P14T, +10.6 on CSL-daily, +6.2 on OpenASL, and +1.3 on How2Sign. It also boosts the R@1 score by +8.3 on P14T, +14.4 on CSL-daily, and +5.9 on How2Sign. Additionally, we set a new baseline for the OpenASL dataset in the SLRet task.
    
\end{abstract}


\section{Introduction}

\begin{figure}[!t]
  \centering
  \includegraphics[width=1
  \linewidth]{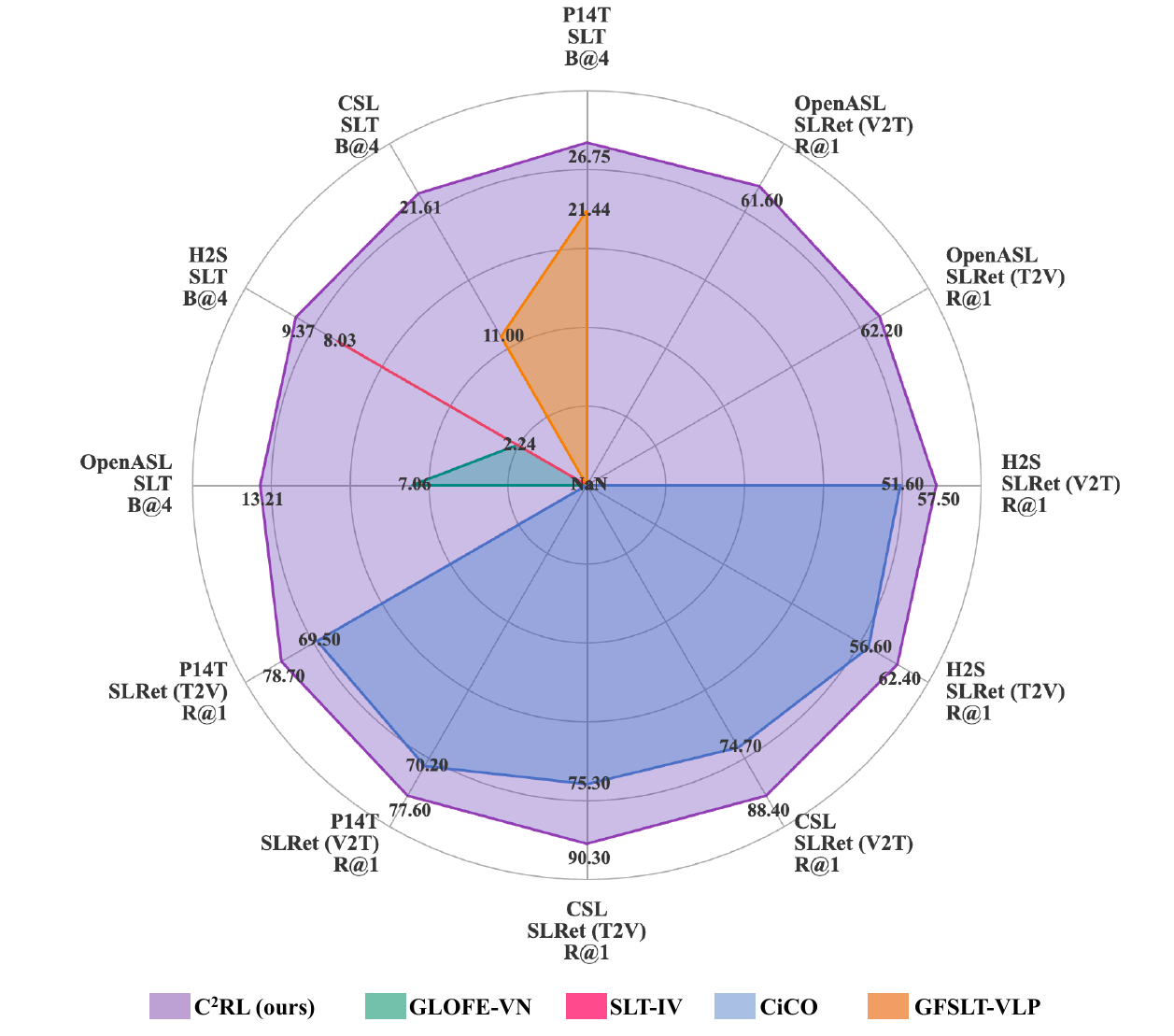}

  \caption{
 Comparative evaluation with gloss-free state-of-the-art methods in Sign Language Translation (SLT) and Sign Language Retrieval (SLRet) tasks, trained only on the corresponding dataset. The proposed C$^2$RL exhibits noteworthy superiority and robust performance across both SLT and SLRet tasks, as evidenced by comprehensive assessments on four diverse datasets.}
  \label{fig:leida}
\end{figure}

\IEEEPARstart{S}{ign} Language Representation Learning (SLRL) plays a pivotal role in sign language understanding, which is essential for a range of sign language-related downstream tasks. Among these tasks, Sign Language Translation (SLT)~\cite{camgoz2020sign} and Sign Language Retrieval (SLRet)~\cite{Duarte_2022_CVPR,cheng2023cico} stand out as key applications. Recent approaches on SLRL mostly focused on two axes: gloss-based and gloss-free. The gloss-based approaches~\cite{camgoz2020sign,chen2022simple,zhou2021spatial,chen2022twostream} rely on an intermediary representation known as \textit{Glosses}\footnote{Glosses are essentially written annotations that provide a rough, sign-to-word transcription of sign language.}. These methods initially capture the literal equivalence of signs and then employ this equivalence as an intermediate form of supervision, to learn locally temporal-aggregated sign language features. However, given that fine-grained gloss annotations are not always available~\cite{Zhou_2023_ICCV,yin2023gloss}, the gloss-free approach has gradually attracted the attention of researchers. 

The gloss-free approaches~\cite{li2020tspnet,zhao2021conditional,yin2023gloss} seek to directly supervise the learning of sign language representations without the intermediary step of glossing, making it highly scalability and greater application potential. 
With this in mind, many exceptional gloss-free approaches have recently been developed.
CiCo~\cite{cheng2023cico} and GFSLT-VLP~\cite{Zhou_2023_ICCV} endeavor to learn language-informative sign representations, drawing inspiration from Vision-Language Pretraining (VLP)~\cite{radford2021learning}, specifically for SLRet and SLT tasks. GloFE-VN~\cite{lin2023gloss} introduces the concept of using conceptual anchors as reference points during translation, thereby enhancing the alignment of visual features with their corresponding textual meanings. Additionally, How2Sign~\cite{duarte2021how2sign} and OpenASL~\cite{shi2022open} contribute by introducing larger and more complex datasets, addressing the critical need for better gloss-free methods. 
Despite promising results, most of these efforts still encounter trivial solutions when optimizing from sign representation learning, attributable to the challenges involved in directly encoding the complex and context-sensitive nature of sign language videos, primarily manifested in the struggle to discern essential sign features through a non-monotonic video-text alignment strategy. As a result, previous methods often suffer from suboptimal performance in associated downstream tasks, as illustrated in Fig.\ref{fig:leida}. 

\begin{figure*}[t]
  \centering
  \includegraphics[width=1.0
  \linewidth]{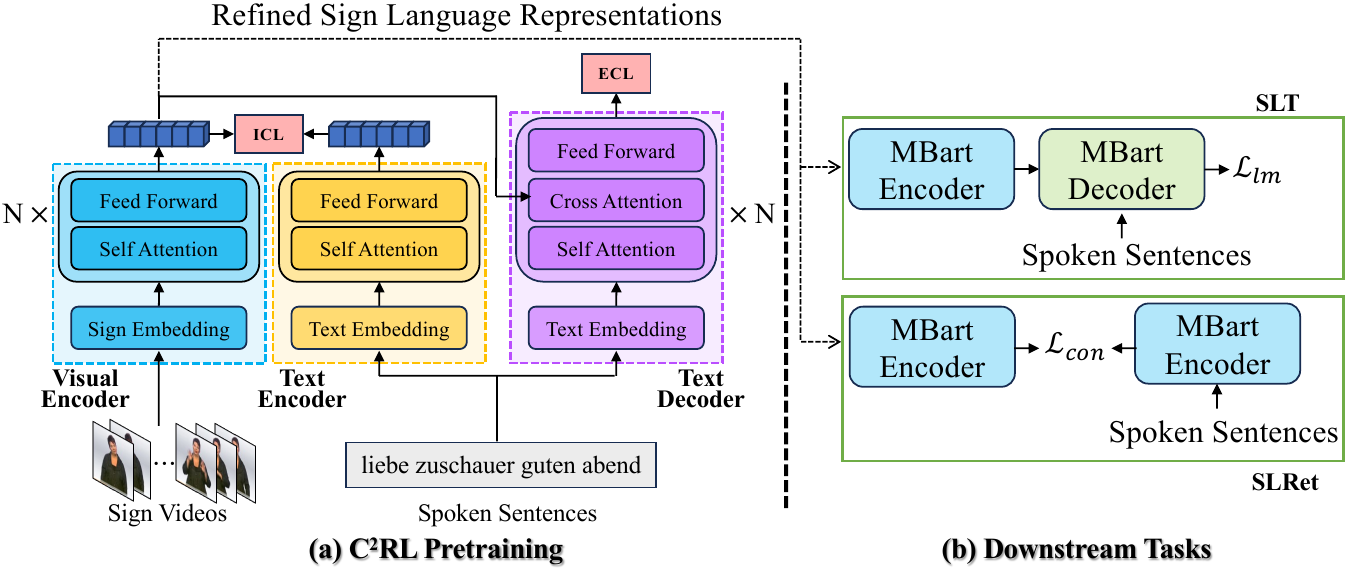}

  \caption{Overview of the proposed framework. C${^2}$RL learns the language-informative visual representation from the perspective of content perception (ICL) and context understanding (ECL). The refined sign language representation is fed into the pretrained language models to adapt effectively to a diverse array of sign language-related downstream tasks.}
  \label{fig:Overview}
\end{figure*}

In this work, we address this challenge by introducing an innovative pretraining paradigm called C${^2}$RL, which integrates Implicit Content Learning (ICL) and Explicit Context Learning (ECL) for universal sign language representation learning, aiming to match the effectiveness of \textit{monotonic alignment}\footnote{Monotonic alignment refers to a one-to-one correspondence between video clips and phrases.} results. As illustrated in Fig.\ref{fig:Overview}, the entire learning process is divided into two distinct stages accordingly pretraining and fine-tuning. Our main contribution lies in the learning of sign language representations in the first stage. In this stage, we elaborately configure two tasks namely ICL and ECL. 

\textbf{In coping with the first aspect}, we draw upon the methodologies outlined in~\cite{cheng2023cico} by incorporating a fine-grained visual-text alignment constraint between the visual and text encoders. More specifically, to learn the implicit contextual nuances of communication, we first extract features from both the vision and linguistic modalities of sign language, and then align these visual elements (like body language) with the corresponding text counterparts (like isolated phrases) in a joint multimodal semantic space. Thereby encouraging the vision encoder to enable a more nuanced understanding of how visual cues in sign language correspond to textual concepts. 
\textbf{In addressing the second aspect}, we draw inspiration from Coca~\cite{yu2022coca} and impose an explicit context constraint between the vision encoder and text decoder. Briefly, this process bears resemblance to a basic translation task, taking visual features and shifted sentences as input to the text decoder and outputting the target spoken language. 
However, the key distinction lies in the purpose of this subtask, which is to compel the vision encoder to thoroughly understand the semantic meanings of the signs and ensure that the vision encoder captures the essential semantic content.

Subsequently, we engage in joint optimization of both the ICL and ECL components to develop unified sign representations. The refined representation then serves as the input for the subsequent pretrained language model, facilitating consistent improvements in gloss-free sign language translation and retrieval tasks.
This work is somewhat inspired by BLIP~\cite{li2022blip}, but fundamentally differs as BLIP focuses on images and lacks the capability to handle temporal dependencies and non-manual features in sign language, which requires specialized modeling techniques and significant adjustments to the model architecture and training approach for effective translation.
Ultimately, as shown in Fig.\ref{fig:leida}, the proposed C${^2}$RL exhibits outstanding performance across four sign language datasets and consistently excels in all major protocols of both SLT and SLRet.

Summary, our contributions to this work can be summarized as follows:
\begin{itemize}
    \item Our study has led to significant advancements in Sign Language Translation (SLT) and Sign Language Retrieval (SLRet) tasks. Compared to existing gloss-free methods, our approach shows remarkable improvements on four major sign language datasets in English, German, and Chinese, and even improves \textbf{$\uparrow$10+} in terms of BLEU-4 score on the CSL-daily dataset. Additionally, the proposed method notably enhances SLRet performance, with average improvements of \textbf{$\uparrow$5+} in terms of R@1 score across these datasets.
    \item We introduce an innovative pretraining paradigm called C${^2}$RL for SLRL, which integrates Implicit Content Learning (ICL) and Explicit Context Learning (ECL) for universal sign language representation learning, thereby achieving visual-text consistent sign language representation learning under non-monotonic alignment conditions.
    \item  C${^2}$RL can be a universal sign language representation extractor, designed for flexible adaptation to a diverse range of downstream tasks related to sign language and offers seamless integration with various off-the-shelf language model backbones.
\end{itemize}

\section{Related Work}
\subsection{Sign Language Understanding}
Sign language understanding encompasses the interpretation and analysis of sign language, with Sign Language Representation Learning (SLRL) being a crucial component. Recently, several key tasks have emerged in this field, including Sign Language Recognition (SLR)~\cite{joze2018ms,huang2018attention,li2020transferring,koller2019weakly,camgoz2020sign,wei2020semantic,yin2023ste,chen2022twostream,zuo2022c2slr}, Sign Language Translation (SLT)~\cite{camgoz2018neural,camgoz2020sign,chen2022simple,li2020tspnet,yin2023gloss,Zhou_2023_ICCV}, and Sign Language Retrieval (SLRet)~\cite{athitsos2010large,Duarte_2022_CVPR,cheng2023cico}. Each of these tasks aims to address different challenges in making sign language more accessible and interpretable. This paper is primarily focused on the tasks of SLT and SLRet.

\paragraph{SLT Stream} Serving as an intermediary step in gloss-based SLT, SLR tasks like Isolated Sign Language Recognition (ISLR)~\cite{joze2018ms,huang2018attention,li2020transferring} and Continuous Sign Language Recognition (CSLR)~\cite{koller2019weakly,camgoz2020sign,wei2020semantic,chen2022twostream,zuo2022c2slr,yin2023ste,zhao2024masa}, primarily tasked with recognizing glosses in sign language videos. It can help in learning temporally aggregated intermediate representations. Therefore, sweetened by this intermediate sign, previous efforts take ISLR or CSLR as a pre-task to improve the SLT. As the representative, \cite{camgoz2018neural} and \cite{camgoz2020sign} introduce end-to-end architectures employing multi-task learning to achieve both CSLR and SLT tasks simultaneously. To further enhance visual perception, researchers have developed powerful visual encoders. For instance, STMC~\cite{zhou2021spatial} proposes a spatial-temporal multi-cue network for both CSLR and SLT, while CorrNet+~\cite{hu2024corrnet+} offers a holistic perspective on human body movements by building fine-grained human body trajectories. Given the scarcity of SLT data, SignBT~\cite{zhou2021improving} back-translates monolingual text into its gloss sequence to generate paired sign sequences, thus providing more data for SLT. Most recently, leveraging the capabilities of pre-trained large language models (LLMs), Chen \etal~\cite{chen2022simple, chen2022twostream} have transferred LLMs to the sign language domain, significantly improving SLT performance. To bridge the modality gap between sign language videos and spoken language text, CV-SLT~\cite{zhao2024conditional} introduces two paths with Kullback-Leibler divergences between the encoder and decoder for cross-modal alignment. Despite their improved performance, \textit{these methods heavily depend on gloss annotations, restricting their generalizability.}

With this in mind, SLT methods have been developed towards the gloss-free direction.
TSPNet~\cite{li2020tspnet} uses a multi-scale representation to address the continuous nature of sign language in videos and explores the use of hierarchical feature learning to learn valid video representations, to reduce the reliance on gloss.
GASLT~\cite{yin2023gloss} develops a dynamic attention mechanism, allowing the model to focus on local semantic units within the video to better understand sign language at the sentence level.
GFSLT-VLP~\cite{Zhou_2023_ICCV} provides a novel pretraining paradigm combining masked self-supervised learning with CLIP, aiming to extract more valid visual representations.
GloFE-VN~\cite{lin2023gloss} innovatively introduces ``conceptual anchors" (common verbs and nouns in spoken language) as references during translation, making the method align visual features with corresponding textual meanings. AVRET~\cite{LIU2024improveing} introduces an adaptive video representation-enhanced Transformer to address temporal correspondence constraints between SL videos and glosses and tackle weakly supervised sequence labeling between long SL videos and sentences. Although achieving its initial success, \textit{these methods are still hindered by rudimentary sign language representations, making them a far cry from their gloss-based counterparts.}

\paragraph{SLRet Stream} SLRet, another crucial area derived from sign language understanding, also grapples with the challenge of suboptimal feature quality. Essentially, SLRet needs to effectively locate and extract relevant sign language content based on queries~\cite{athitsos2010large}. This typically involves using sign language representations to match queries while also requiring an understanding of context to ensure the relevance and accuracy of retrieval results. Currently, there are still relatively few efforts in this direction. As the pioneering work, SPOT-ALIGN~\cite{Duarte_2022_CVPR} addresses this challenge by formulating sign language retrieval purely as a video-text retrieval task, modeling cross-modal alignment based on the overall global embeddings of sign language videos and text. Building on this approach, Cico~\cite{cheng2023cico} attempts to introduce the Vision-Language Pretraining (VLP) schema like CLIP~\cite{radford2021learning} into sign language retrieval.

All of this work relies heavily on informative sign language representations. In this study, we enhance the semantic representation of sign features by learning universal concepts, recognizing explicit context, and understanding implicit content. This dual approach is crucial for accurately capturing and conveying the intent and nuances of the original sign message.

\subsection{Visual-Language Pretraining} 

This section reviews Vision-Language Pretraining (VLP), as this work drew partial inspiration from this framework. VLP trains models to process and understand multimodal information by using large datasets that contain both visual (images or videos) and textual (sentences or captions) data. As one of the most influential works, CLIP~\cite{radford2021learning}, focusing on zero-shot and few-shot learning, significantly narrows the gap between vision and language feature representation by learning from vast amounts of image-text pairs scraped from the internet. Subsequently, BLIP~\cite{li2022blip} and BLIP V2\cite{li2023blip} build on this foundation by introducing a bootstrapped image-language pretraining framework that leverages both synthetic and real image-caption pairs. In parallel, pertaining paradiums on multimodal understanding and reasoning~\cite{lu2019vilbert,chen2020uniter}, image-to-text retrieval~\cite{jia2021scaling}, image captioning~\cite{vlp1}, video captioning~\cite{vlp3,vlp5} as well as general VLP ~\cite{li2022blip,li2019visualbert,vlp2,vlp4} have been proposed over the years. Sign Language Representation Learning (SLRL), inherently aligned with its task-specific characteristics, is ideally positioned to leverage VLP. 
However, applying the VLP in sign language analysis tasks is not straightforward, especially due to the non-monotonic semantic alignment between sign videos and corresponding textual, making it difficult to learn a visual-text-consistent sign representation. As we have proven, the proposed C$^2$RL can significantly enhance feature quality and achieve high performance through comprehensive constraints on both content and context.

\section{Method}
In this section, we begin by presenting the overview of the proposed framework in Section \ref{sec:framework}. Then, we detail the objectives of pretraining in Section \ref{sec:pretraining}. Lastly, we illustrate the application of our learned video representations of sign language in specific downstream tasks in Section \ref{sec:downstream}. 

\subsection{Training Framework}\label{sec:framework}
The proposed C${^2}$RL framework consists of a \textit{visual encoder} for processing sign language videos, a \textit{text encoder} for extracting text information, and a \textit{text decoder} for generating spoken sentences. Additionally, when conducting downstream tasks, we cascade the off-the-shelf pretrained \textit{MBart}~\cite{liu2020multilingual} to harvest its advanced generative language capabilities. The whole architecture is illustrated in Fig.\ref{fig:Overview}.

\paragraph{Visual Encoder} The visual encoder comprises a sign embedding layer and transformer encoder blocks. Unlike text inputs which can be directly transformed into continuous vectors by tokenizer and text embedding, sign language videos also contain spatial information. Therefore we constructed the sign embedding module as shown in Table~\ref{tab:model}. A batch sign language videos $V=(v_1,v_2...v_N)$ with a size of $(N \times T \times 224 \times 224 \times 3)$ is first passed through a ResNet~\cite{he2016deep} pretrained on ImageNet~\cite{deng2009imagenet} to extract the spatial features of each frame. We utilize a combination of Conv1D-BN1D-ReLU modules to model the local temporal information between adjacent frames. Then a fully connected layer aligns the dimensions to the hidden size of the transformer blocks. The transformer encoder blocks are used to capture the global context relationship of the input.
\begin{table}[t]
    \centering
    \caption{Detailed sign embedding framework. N means batch size. T means the lengths of the longest input sign video in the batch. C means the hidden size of the following transformer blocks.
    }
    \begin{tabular}{c|c}
    \toprule
    
    Module & Output Size\\
    \midrule
    Sign video Input   & $N \times T \times 224 \times 224 \times 3$ \\ 
    Resnet wo/ FC     & $N \times T \times 512$\\
    Conv1D-BN1D-ReLU   & $N \times T \times 512$\\
    Fully Connected   & $N \times T \times C$\\
    \bottomrule
    \end{tabular}
    \label{tab:model}
\end{table}

\paragraph{Text Encoder} The text encoder consists of a text embedding layer and transformer encoder blocks. Given a batch of spoken sentences $T=(t_1,t_2...t_N)$, the text embedding layer that is a simple matrix maps it into vectors. The transformer encoder blocks capture the global context information and extract the features of input sentences.

\paragraph{Text Decoder} The text decoder consists of a text embedding layer and transformer decoder blocks. The transformer decoder blocks utilize cross-attention to fuse sign language features and text features. The whole sentence is then generated token by token in an autoregressive manner.

\paragraph{MBart} As shown in Fig.~\ref{fig:Overview}, we choose a pretrained language model with the encoder-decoder architecture since it has both language understanding and text generation capabilities. Following~\cite{chen2022simple,chen2022twostream}, we use MBart~\cite{liu2020multilingual} as our model. MBart is a sequence-to-sequence denoising auto-encoder pretrained on extensive monolingual corpora across multiple languages. It has a standard transformer~\cite{vaswani2017attention} architecture with 12 layers of the encoder and 12 layers of the decoder. Originally designed for translation purposes, MBart has demonstrated notable enhancements in the performance of SLT~\cite{chen2022simple}. 

\subsection{C${^2}$RL Pretraining}\label{sec:pretraining}
During C${^2}$RL pretraining, we jointly optimize two objectives: Implicit Content Learning (ICL) through contrastive loss and  Explicit Context Learning (ECL) through language modeling loss. 

\paragraph{Implicit Content Learning (ICL)} This objective aims to learn the implicit contextual nuances of communication from the vision and linguistic modalities of sign language, to achieve a more nuanced understanding of how visual cues in sign language correspond to textual concepts. It can bring the positive pairs closer together and pull the negative pairs further apart by contrasting the visual linguistic similarity matrix.
We achieve this goal by calculating the similarity matrix using global features like CLIP~\cite{radford2021learning} as it has been proven effective on many visual language tasks. 
However, since sign language videos carry more semantics, a more nuanced similarity computation is needed. Therefore, we adopt Cross-Lingual Contrastive Learning (CLCL) proposed by~\cite{cheng2023cico} to get the Video-to-Text similarity matrix $Z_{V2T}$ and Text-to-Video similarity matrix $Z_{T2V}$. CLCL contrasts the
sign videos and corresponding text transcripts in a joint embedding space, and concurrently pinpoints fine-grained cross-lingual (sign-to-word) connections between two language types. Then we utilize InfoNCE loss~\cite{gutmann2010noise} to compute the contrastive loss:
\begin{equation}
\label{eq:con}
\begin{split}
    \mathcal{L}_{con}  = &-\frac{1}{N}\sum_{i=1}^{N}\log\frac{\exp{(Z^{(i,i)}_{V2T}/\tau)}}{\sum_{j=1}^{N}\exp{(Z^{(i,j)}_{V2T}/\tau)}} \\ 
      & -\frac{1}{N}\sum_{i=1}^{N}\log\frac{\exp{(Z^{(i,i)}_{T2V}/\tau)}}{\sum_{j=1}^{N}\exp{(Z^{(i,j)}_{T2V}/\tau)}} 
\end{split}
\end{equation}
where $Z^{(i,i)}_{V2T}$ denotes the similarity of $(v_i,t_j)$ and $\tau$ is a trainable temperature parameter.

\paragraph{Explicit Context Learning (ECL)} This objective aims to train the model to generate spoken sentences token by token autoregressively based on the given sign language representation in detailed granularity. By optimizing the language modeling (LM) loss, the text decoder learns to maximize the conditional likelihood of the video-text pairs $(v_i,t_i)$ which can be formulated as follows:
\begin{equation}
\label{eq:lm}
\mathcal{L}_{lm} = -\frac{1}{N}\sum_{i=1}^{N} \sum_{u=1}^{U} \log p(t^u_i|t^{<u}_i,v_i)
\end{equation}
where $t^u_i$ represents the $u$-th token of sentence $t_i$.

\paragraph{Loss Function} We optimize the entire model framework using a weighted sum of contrast loss and language modeling loss:
\begin{equation}
\label{eq:all}
\mathcal{L} = \alpha \mathcal{L}_{con} + \beta \mathcal{L}_{lm}
\end{equation}
where $\alpha$ and $\beta$ are two weight coefficients, whose impact on performance we study in Section.\ref{exp:abl}.

\subsection{Downstream Tasks}\label{sec:downstream}
After the joint optimization of the ICL and the ECL, we achieve a robust visual encoder endowed with strong sign language feature modeling capabilities. This encoder is then employed as a universal sign language representations extractor for the following downstream tasks, including Sign Language Translation (SLT) and Sign Language Retrieval (SLRet) tasks.

\paragraph{Sign Language Translation (SLT)} This task aims to convert sign language representations into corresponding spoken sentences. Specifically, we first use a fully connected (FC) layer to align the dimensions of the sign language representation with the dimensions of the MBart. Then, the sign language representation is sequentially passed through the layers of the MBart encoder to get hidden states. After that, the MBart decoder takes shifted sentences and the hidden states as inputs for output prediction. We optimize the SLT model using language modeling loss as illustrated in equation~\ref{eq:lm}.

\paragraph{Sign Language Retrieval (SLRet)} 
This task focuses on identifying corresponding pairs of sign language videos and spoken text within the dataset. It can be divided into two sub-tasks: text-to-video (T2V) and video-to-text (T2V). Similar to SLT finetuning, the sign language representations are first transformed into the same dimensions as MBart through an FC layer. Subsequently, one MBart encoder is employed to extract the hidden states of these representations, while another MBart encoder is tasked with extracting the hidden states of the corresponding text transcripts. These two encoders operate independently, without shared parameters, to accommodate the distinct input modalities. Finally, we compute the similarity matrix using CLCL proposed by \cite{cheng2023cico} and train the SLRet model using the contrastive loss illustrated in equation~\ref{eq:con}.

\section{Experiments}
In this section, we first provide details on the benchmark datasets and the experimental setup. Then, we thoroughly evaluate the impact of C$^2$RL on Sign Language Translation (SLT) and Sign Language Retrieval (SLRet) tasks, comparing our results with state-of-the-art methods across four benchmark datasets. Finally, we conduct extensive ablation studies to explore the effectiveness of each component in C$^2$RL.

\subsection{Datasets and Metrics}
\paragraph{Datasets} We evaluate our proposed method on four mainstream datasets: PHOENIX-2014T~\cite{camgoz2018neural}, CSL-Daily~\cite{zhou2021improving}, How2Sign~\cite{duarte2021how2sign}, and OpenASL~\cite{shi2022open}.
The \textbf{PHOENIX-2014T dataset}~\cite{camgoz2018neural} is a German sign language (DGS) dataset taken from a TV broadcast whose topic focuses on weather forecasts. It contains 8,257 sign-video-German-text pairs divided into train, validation, and test with a size of 7,098, 519, and 642, respectively. 
The \textbf{CSL-Daily dataset}~\cite{zhou2021improving} is a Chinese sign language (CSL) dataset recorded in the laboratory whose topic focuses on daily life. It contains 20,654 sign-video-Chinese-text pairs divided into train, validation, and test with a size of 18,401, 1,077, and 1,176, respectively.
The \textbf{How2Sign dataset}~\cite{duarte2021how2sign} is an American sign language (ASL) dataset recorded in the laboratory that focuses on instructional topics corresponding to various categories. It contains 35,260 sign-video-English-text pairs divided into train, validation, and test with a size of 31,164, 1,740, and 2,356, respectively. Following~\cite{cheng2023cico}, we eliminated the invalid pairs where the subtitle alignment was found to exceed the duration of the video, left with 31,085, 1,739, and 2,348 available pairs in the train, validation, and test sets, respectively. We also use Fast R-CNN to crop videos, keeping only the part of the signer.
The \textbf{OpenASL dataset}~\cite{shi2022open} is an American sign language (ASL) dataset collected from online videos that cover various topics. It contains 98,417 sign-video-English-text pairs divided into train, validation, and test with a size of 96,476, 966, and 975, respectively.


\paragraph{Evaluation Metrics} The \textbf{SLT task} involves directly converting sign language videos into spoken sentences, resembling a standard translation task commonly found in NLP. Consequently, following previous work~\cite{camgoz2018neural,Zhou_2023_ICCV}, we use BLEU~\cite{papineni2002bleu} and ROUGE~\cite{lin2004rouge} to evaluate performance in this task. The \textbf{SLRet task}, on the other hand, involves retrieving corresponding sign videos from the test set given spoken texts (T2V) or retrieving corresponding texts given videos (V2T), making it akin to a video-text retrieval task. For this task, recall at rank K (R@K) is used to evaluate retrieval performance, as per~\cite{cheng2023cico}. \textit{Higher values} indicate better performance for all metrics. Here, all results are reported on the test set.


\subsection{Implementation Details}


\paragraph{Data Pre-processing} For video pre-processing, we downsample the input sign video with a rate of $k$. The video with $T$ frames is split into $k\times T$ clips sequentially. \textbf{During training}, we randomly choose one frame in each clip and compose them into a new video. \textbf{During inference}, we choose the first frame in each clip. Unless otherwise specified, we use $k=25\%$ for the downsampling rate. Each frame in the input video are first resized to $256 \times 256$ and then randomly/centerly cropped into $224 \times 224$ during the training/inference stage. For text pre-processing, we utilize the standard MBart tokenizer~\cite{liu2020multilingual} to split the spoken sentences into smaller units (tokens) and map them into their corresponding IDs in the vocabulary.

\paragraph{C$^2$RL Pretraining} 
We use ResNet-18~\cite{he2016deep}, pretrained on ImageNet~\cite{deng2009imagenet}, as the backbone for sign embedding mapping. The fully connected layer at the end of ResNet-18 is replaced with a temporal module, which consists of Conv1d-BN-ReLU layers. The Conv1d layer has a kernel size of 3 and a stride of 1. The transformer blocks used in this stage have 3 layers for both the encoder and decoder, with each layer having 8 attention heads, a hidden size of 512, and a feed-forward dimension of 2048. Dropout is set to 0.1 to prevent overfitting. During training, we use the SGD optimizer~\cite{robbins1951stochastic} with a momentum of 0.9. The learning rate is initialized at $1 \times 10^{-2}$ and decays to 0 using a cosine annealing scheduler~\cite{loshchilov2016sgdr}. The model is trained on 8 NVIDIA GeForce RTX 3090 GPUs with a batch size of 8 per process. We employ cross-entropy loss with a label smoothing of 0.2, training the model for 200 epochs.

\begin{table}[t]
  \centering
   \caption{SLT results on PHOENIX14T dataset. $*$ denotes methods reproduced by \protect\cite{yin2023gloss}. We \textbf{bold} the best results in the gloss-based setting and gloss-free setting. Improvement represents comparisons with the previous best gloss-free result.}
  \resizebox{\linewidth}{!}{
  \begin{tabular}{@{}l|ccccc@{}}
  \toprule
    { Method} &   Rouge &  B@1 &  B@2 &  B@3 &  B@4 \\
    \midrule
     \rowcolor[gray]{.8} \multicolumn{6}{c}{Gloss-based}\\
     \midrule
     SLRT~\cite{camgoz2020sign} & - & 46.61 & 33.73 & 26.19 & 21.32 \\
     SignBT~\cite{zhou2021improving} & 49.54 & 50.80 & 37.75 & 29.72 & 24.32 \\
     MMTLB~\cite{chen2022simple}  & 52.65 & 53.97 & 41.75 & 33.84 & 28.39 \\
     SLTUNET~\cite{zhang2023sltunet} & 52.11 & 52.92 & 41.76 & 33.99 & 28.47 \\
     TS-SLT~\cite{chen2022twostream}&  53.48 &  \textbf{54.90} & 42.43 & 34.46 & 28.95 \\
     CV-SLT~\cite{zhao2024conditional}&  \textbf{54.33} &  54.88 & \textbf{42.68} & \textbf{34.79} & \textbf{29.27} \\
     \midrule
     \rowcolor[gray]{.8} \multicolumn{6}{c}{Gloss-free}\\
     \midrule
     NSLT~\cite{camgoz2018neural} & 31.80 & 32.24 & 19.03 & 12.83 & 9.58  \\
     SLRT-GF$^*$~\cite{camgoz2020sign}  & 31.10 & 30.88 & 18.57 & 13.12 & 10.19 \\
     TSPNet~\cite{li2020tspnet} &  34.96 & 36.10  & 23.12 & 16.88 & 13.41  \\
     CSGCR~\cite{zhao2021conditional} & 38.85 & 36.71 & 25.40 & 18.86 & 15.18  \\
     GASLT~\cite{yin2023gloss} & 39.86 & 39.07  & 26.74 & 21.86 & 15.74  \\
    GFSLT-VLP~\cite{Zhou_2023_ICCV}  &\ 42.49 & 43.71 & 33.18 & 26.11 & 21.44 \\
     \midrule
      \textbf{C$^2$RL(ours)} &  \textbf{50.96} & \textbf{52.81} & \textbf{40.20} & \textbf{32.20} & \textbf{26.75}\\
      Improvement  & +8.47 & +9.10 & +7.02 & +6.09& +5.31\\
    \bottomrule
  \end{tabular}}
  \label{tab:PHOENIX14T}
\end{table}

\paragraph{Dowmstream Tasks} The offline sign features extracted by the learned visual encoder are used as the input for MBart. We initialize MBart weights with the official release of MBart-large-cc25\footnote{\url{https://huggingface.co/facebook/MBart-large-cc25}}. For memory efficiency, the fully connected layer and corresponding tokenizer are trimmed using the translation corpus of the target training set. Dropout is set to 0.3 to prevent overfitting.
During training, we use the Adam optimizer~\cite{kingma2014adam}. \textbf{For SLT}, the learning rate is initialized at $1 \times 10^{-5}$ for MBart, and $1 \times 10^{-3}$ for the projection layers before MBart. Both learning rates decay to 0 using a cosine annealing scheduler~\cite{loshchilov2016sgdr}. \textbf{For SLRet}, the learning rate is initialized at $1 \times 10^{-4}$ for all weights in the model and decayed using the same scheduler used in SLT. The model is trained on 8 NVIDIA GeForce RTX 3090 GPUs, with each process having a batch size of 16, and the training lasts for 80 epochs.


\begin{table}[t]
  \centering
  \caption{SLT results on CSL-daily dataset. $*$ denotes methods reproduced by \protect\cite{yin2023gloss}. $\dag$ denotes methods reproduced by \protect\cite{zhou2021improving}. 
  }
  \resizebox{\linewidth}{!}{
  \begin{tabular}{@{}l|ccccc@{}}
    \toprule
    { Method}  &   Rouge&  B@1 &  B@2 &  B@3 &  B@4 \\
    \midrule
     \rowcolor[gray]{.8} \multicolumn{6}{c}{Gloss-based}\\
     \midrule
    
     SLRT$^\dag$~\cite{camgoz2020sign}   & 36.74 & 37.38 & 24.36 & 16.55 & 11.79 \\
     SignBT~\cite{zhou2021improving}  & 49.31 & 51.42 & 37.26 & 27.76 & 21.34 \\
    MMTLB~\cite{chen2022simple}  & 53.25 & 53.31 & 40.41 & 30.87 & 23.92 \\
    SLTUNET~\cite{zhang2023sltunet} & 54.08 & 54.98 & 41.44 & 31.84 & 25.01 \\
     TS-SLT~\cite{chen2022twostream}&  55.72 &  55.44& 42.59 & 32.87 & 25.79 \\
     CV-SLT~\cite{zhao2024conditional}&  \textbf{57.06} &  \textbf{58.29} & \textbf{45.15} & \textbf{35.77} & \textbf{28.94} \\
    \midrule
     \rowcolor[gray]{.8} \multicolumn{6}{c}{Gloss-free}\\
     \midrule
     NSLT$^\dag$~\cite{camgoz2018neural}  & 34.54 & 34.16 & 19.57 & 11.84 & 7.56  \\
     TSPNet$^*$~\cite{li2020tspnet} & 18.38 & 17.09  & 8.98 & 5.07 & 2.97  \\
     GASLT~\cite{yin2023gloss}  &  20.35 & 19.90  & 9.94& 5.98 & 4.07  \\
     GFSLT-VLP~\cite{Zhou_2023_ICCV}   & 36.44 & 39.37 & 24.93 & 16.26 & 11.00 \\
     \midrule
      \textbf{C$^2$RL(ours)} & \textbf{48.21} & \textbf{49.32} & \textbf{36.28} & \textbf{27.54} & \textbf{21.61}\\
      Improvement  & +11.77 & +10.05 & +11.35& +11.28& +10.61\\
    \bottomrule
  \end{tabular}}
  \label{tab:CSL-Daily}
\end{table}

   
    
    


    

\begin{table}[t]
  \centering
  \caption{SLT results on How2Sign dataset. We \textbf{bold} the best results obtained after pretraining with/without auxiliary data. Improvement represents comparisons with the previous best result without using auxiliary data. 
  \C means pretrained with additional sign language data (YT-ASL\protect\cite{uthus2023youtubeasl}) and vice versa.
  YT-ASL is approximately 20$\times$ times larger than How2Sign.
  }
  \resizebox{\linewidth}{!}{
  \begin{tabular}{@{}l|c|ccccc@{}}
   
    \toprule
    { Method}   & YT-ASL &  Rouge &  B@1 &  B@2 &  B@3 &  B@4 \\
    \midrule
     \rowcolor[gray]{.8} \multicolumn{7}{c}{Gloss-free}\\
     \midrule
     TF-H2S~\cite{alvarezsign} & \X & - & 17.40 & 7.69 & 3.97 & 2.21  \\
     GloFE-VN~\cite{lin2023gloss} & \X & 12.61 & 14.94  & 7.27 & 3.93 & 2.24 \\
     SLT-IV~\cite{tarres2023sign} & \X & - & \textbf{34.01}  & \textbf{19.30} & 12.18 & 8.03  \\
     T5-SLT ~\cite{uthus2023youtubeasl} & \X & - & 14.96  & 5.11 & 2.26 & 1.22 \\
      \midrule
      \textbf{C$^2$RL(ours)}  & \X & \textbf{27.02} & 29.07 & 18.56 & \textbf{12.92} & \textbf{9.37}\\
      Improvement &  & +14.41 & -4.94 & -0.74 & +0.47& +1.34\\
       \midrule
    \textcolor{gray!60}{T5-SLT ~\cite{uthus2023youtubeasl}}  & \C & - & \textcolor{gray!60}{37.82}  & \textcolor{gray!60}{24.13} & \textcolor{gray!60}{16.92} & \textcolor{gray!60}{12.39} \\
     \bottomrule
    
  \end{tabular}}
  \label{tab:How2Sign}
\end{table}

\begin{table}[t]
  \centering
   \caption{SLT results on OpenASL dataset. $\dag$ denotes methods reproduced by \protect\cite{shi2022open}. 
  }
  \resizebox{\linewidth}{!}{
  \begin{tabular}{@{}l|ccccc@{}}
   
    \toprule
    { Method} &   Rouge &  B@1 &  B@2 &  B@3 &  B@4 \\
     \midrule
     \rowcolor[gray]{.8} \multicolumn{6}{c}{Gloss-free}\\
     \midrule
     Conv-GRU$^\dag$~\cite{camgoz2018neural}  & 16.10 & 16.11 & 8.85 & 6.18 & 4.58 \\
     I3D-transformer~\cite{shi2022open} & 18.64 & 18.31 & 10.15 & 7.19 & 5.66 \\
     OpenASL~\cite{shi2022open}  & 21.02 & 20.92 & 12.08 & 8.59 & 6.72  \\
     GloFE-VN~\cite{lin2023gloss} & 21.75 & 21.56  & 12.74 & 9.05 & 7.06 \\
    \midrule
      \textbf{C$^2$RL(ours)} & \textbf{31.36} & \textbf{31.46} & \textbf{21.85} & \textbf{16.58} & \textbf{13.21}\\
      Improvement  & +9.61& +9.90 & +9.11 & +7.53& +6.15\\
    \bottomrule
    
  \end{tabular}}
  \label{tab:OpenASL}
\end{table}

\subsection{Comparison with State-of-the-Art Methods}

\paragraph{Evaluation on Sign Language Translation (SLT)} 
We evaluate our models on sign language translation across four benchmark datasets. Our method operates in a gloss-free stream, meaning that gloss annotations are not involved in our entire training and inference process. However, for PHOENIX14T and CSL-Daily, which provide gloss annotations, we also make comparisons under the gloss-based setting for these datasets.

\noindent \textbf{Comparison on PHOENIX14T Dataset} \quad As shown in Table~\ref{tab:PHOENIX14T}, the proposed C$^2$RL significantly outperforms the state-of-the-art methods GASLT~\cite{yin2023gloss} and GFSLT-VLP~\cite{Zhou_2023_ICCV} under the gloss-free setting. Specifically, it achieves a notable improvement of \textbf{$\uparrow$5.3} over GFSLT-VLP and \textbf{$\uparrow$11.0} over GASLT in terms of the B@4 score. This represents an average performance increase of \textbf{$\uparrow$25\%}. Furthermore, C$^2$RL narrows the gap with gloss-based methods, reducing the performance gap on B@4 from 7.83 to 2.52, a \textbf{$\downarrow$67.8\%} reduction. This demonstrates that our method can achieve comparable results to gloss-supervised methods, achieving sign language feature understanding at the word granularity level.

\noindent \textbf{Comparison on CSL-Daily Dataset} \quad CSL-Daily is an extremely challenging Chinese dataset. As shown in Table~\ref{tab:CSL-Daily}, most gloss-free methods exhibit very low performance, with B@4 scores typically around $\pm$10. Our method, for the first time, raises the performance of gloss-free methods above 21.6 on this dataset. This is an improvement of \textbf{$\uparrow$10} points in all metrics compared to GFSLT-VLP~\cite{Zhou_2023_ICCV}, and it narrows the performance gap with gloss-based methods from 17.9 to 7.3, a \textbf{$\downarrow$59.2\%} reduction. This substantial improvement suggests that our approach significantly enhances the representation of sign language features, even in the absence of gloss annotations, bringing gloss-free methods closer to the effectiveness of gloss-based methods and demonstrating the robustness and generalizability of C$^2$RL in challenging settings.

\noindent \textbf{Comparison on How2Sign and OpenASL Datasets} \quad How2Sign and OpenASL are two large American Sign Language datasets without gloss annotations. On these datasets, our method demonstrates strong generalization properties. Specifically, without the assistance of additional training data, we achieved a \textbf{$\uparrow$1.34} improvement in B@4 on How2Sign as shown in ~\ref{tab:How2Sign}.  Notably, when compared to T5-SLT~\cite{uthus2023youtubeasl}, which utilizes approximately 20$\times$ more additional data from YT-ASL for pretraining, our results remain promising, trailing only by $\downarrow$3 in B@4. Furthermore, when compared to SLT-IV~\cite{tarres2023sign}, a method specifically designed for the How2Sign dataset, we lag slightly in B@1 and B@2 but show improvements in B@3 and B@4.
Additionally, C$^2$RL performs exceedingly well on the OpenASL dataset as illustrated in Table~\ref{tab:OpenASL}. When compared with the recent method GloFE-VN~\cite{lin2023gloss}, we achieve a \textbf{$\uparrow$6.15} improvement in B@4, marking an average improvement of \textbf{$\uparrow$66\%} across all metrics. This significant enhancement underscores the effectiveness and robustness of our method in handling diverse datasets and showcases its potential in improving sign language translation tasks.

\paragraph{Evaluation on Sign Language Retrieval (SLRet)} Table~\ref{tab:open-asl SLRet} presents a thorough comparison with previous gloss-free SLRet methods across four sign language datasets. Sign language retrieval is a relatively new exploration direction in sign language analysis, with few related works. SA~\cite{Duarte_2022_CVPR} and Cico~\cite{cheng2023cico} are two of the representative methods in this domain. As shown in the table, C$^2$RL substantially outperforms these methods in SLRet tasks. Specifically, in terms of  R@1 score, C$^2$RL achieves significant improvements of \textbf{$\uparrow$9.2} T2V and \textbf{$\uparrow$7.4} V2T on PHOENIX14T, \textbf{$\uparrow$15.0} T2V and \textbf{$\uparrow$13.7} V2T on CSL-Daily, as well as \textbf{$\uparrow$5.8} T2V and \textbf{$\uparrow$5.9} V2T on How2Sign. Additionally, we establish the SLRet baseline performance on the OpenASL dataset, achieving scores of 62.6 T2V and 61.6 V2T at R@1. This high proficiency can be attributed to the holistic impact generated by the synergistic learning of both contextual and content semantics.

\begin{table}[!t]
  \centering
   \caption{SLRet results on four datasets.}
  \resizebox{\linewidth}{!}{
  \begin{tabular}{@{}l|ccc|ccc@{}}
   
    \toprule
    { \multirow{2}{*}{\textbf{Method}} } & \multicolumn{3}{c|}{\textbf{T2V}}& \multicolumn{3}{c}{\textbf{V2T}}\\
    & R@1  &  R@5 &  R@10 & R@1  &  R@5 &  R@10 \\
    \midrule
     \rowcolor[gray]{.8} \multicolumn{7}{c}{PHOENIX14T}\\
     \midrule
     SA~\cite{Duarte_2022_CVPR}& 55.8 & 79.6 & 87.2 & 53.1 & 79.4 & 86.1 \\
    
    CiCo~\cite{cheng2023cico} & 69.5 & 86.6 & 92.1 & 70.2 & 88.0 & 92.8 \\
    \midrule
      \textbf{C$^2$RL(ours)} & \textbf{78.7}  & \textbf{92.2} & \textbf{94.9} & \textbf{77.6} & \textbf{91.3} & \textbf{94.2}\\
      Improvement & +9.2 & +5.6 & +2.8 & +7.4& +3.3& +1.4\\
      \midrule
     \rowcolor[gray]{.8} \multicolumn{7}{c}{CSL-Daily}\\
     \midrule
    CiCo~\cite{cheng2023cico} & 75.3 & 88.2 & 91.9 & 74.7 & 89.4 & 92.2 \\
    \midrule
      \textbf{C$^2$RL(ours)} & \textbf{90.3}  & \textbf{96.4} & \textbf{97.7} & \textbf{88.4} & \textbf{95.7} & \textbf{97.1}\\
      Improvement &+15.0 & +8.2 & +5.8 & +13.7& +6.3& +4.9\\
      \midrule
     \rowcolor[gray]{.8} \multicolumn{7}{c}{How2Sign}\\
     \midrule
    
    SA~\cite{Duarte_2022_CVPR}& 34.2 & 48.0 & 52.6 & 23.6 & 47.0 & 53.0 \\

    CiCo~\cite{cheng2023cico} & 56.6 & 69.9 & 74.7 & 51.6 & 64.8 & 70.1 \\
    \midrule
      \textbf{C$^2$RL(ours)} & \textbf{62.4}  & \textbf{75.9} & \textbf{80.1} & \textbf{57.5} & \textbf{68.4} & \textbf{73.0}\\
      Improvement& +5.8 &+6.0 & +5.4 & +5.9 & +3.6 & +2.9\\
      \midrule
    \rowcolor[gray]{.8} \multicolumn{7}{c}{OpenASL}\\
     \midrule
    
      \textbf{C$^2$RL(ours)} & \textbf{62.2}  & \textbf{81.7} & \textbf{86.8} & \textbf{61.6} & \textbf{79.8} & \textbf{84.6}\\
    \bottomrule
    
  \end{tabular}}
  \label{tab:open-asl SLRet}
\end{table}

\subsection{Ablation Studies} \label{exp:abl}
In this part, we thoroughly investigate the contribution of each component to C$^2$RL.
All ablation studies are evaluated on the test set of the PHOENIX-2014T dataset, as it is a most widely used sign language dataset.

\begin{table}[!t]
    \centering
    \caption{Effect of proposed learning objectives.}
    \resizebox{\linewidth}{!}{
    \begin{tabular}{l|ccc}
    \toprule
    { \multirow{2}{*}{\textbf{Objectives}} } & {\textbf{SLT}}& {\textbf{SLRet (T2V)}}&{\textbf{SLRet (V2T)}}\\
    & B@4  &  R@1 &  R@1  \\
    \midrule
    ICL only & 24.83 & 74.7 & 73.5\\
    ECL only & 25.72 & 74.1 & 74.5 \\
    ICL + ECL & \textbf{26.75} & \textbf{78.7} & \textbf{77.6}\\
    \bottomrule
    \end{tabular}
    }
    \label{tab:Strategy Portfolio}
\end{table}
\paragraph{ICL \& ECL} The most straightforward approach to validate the effectiveness of C$^2$RL is to assess the impact of the two proposed learning objectives, namely Implicit Content Learning (ICL) and Explicit Context Learning (ECL), on downstream results. In this experiment, we first train a visual encoder using these two constraints, either separately or jointly, and then use the trained visual encoder as a sign language feature extractor for downstream tasks. As depicted in Table~\ref{tab:Strategy Portfolio}, co-optimizing ICL and ECL resulted in significant improvements compared to their individual usage. In addition, we found that ECL achieved better results than ICL in SLT (25.72 $vs.$ 24.83 ) and SLRet(V2T) (74.5 $vs.$ 73.5 ), suggesting that understanding the semantic meanings of the signs is more crucial for video-to-text mapping than accurately matching visual cues with corresponding textual concepts. 
Additionally, we also investigated the weight of ICL $\alpha$ and ECL $\beta$ during C$^2$RL pertaining. As illustrated in Table~\ref{tab:weight}, changing the weights of the losses doesn't have much impact on the final performance. The setting of $\alpha : \beta = 1 : 1$ obtains the best results, illustrating that both losses are equally important. Therefore, we set them as 1.0 for the rest of our experiments.

\begin{table}[!t]
    \centering
    \caption{Effect of loss weight.}
    \begin{tabular}{c|ccc}
    \toprule
    { \multirow{2}{*}{ICL($\alpha$) : ECL($\beta$)} } & {\textbf{SLT}}& {\textbf{SLRet (T2V)}}&{\textbf{SLRet (V2T)}}\\
    & B@4  &  R@1 &  R@1  \\
    \midrule
   $1 : 2$ & 26.19 & 77.9 & 77.4\\
    $2 : 1$ & 25.84 & 76.2 & 76.6 \\
    $1 : 1$ & \textbf{26.75} & \textbf{78.7} & \textbf{77.6}\\
    \bottomrule
    \end{tabular}
    \label{tab:weight}
\end{table}

\begin{figure}[!t]
  \centering
   \subfloat[SLT]{\includegraphics[width=0.45\linewidth]{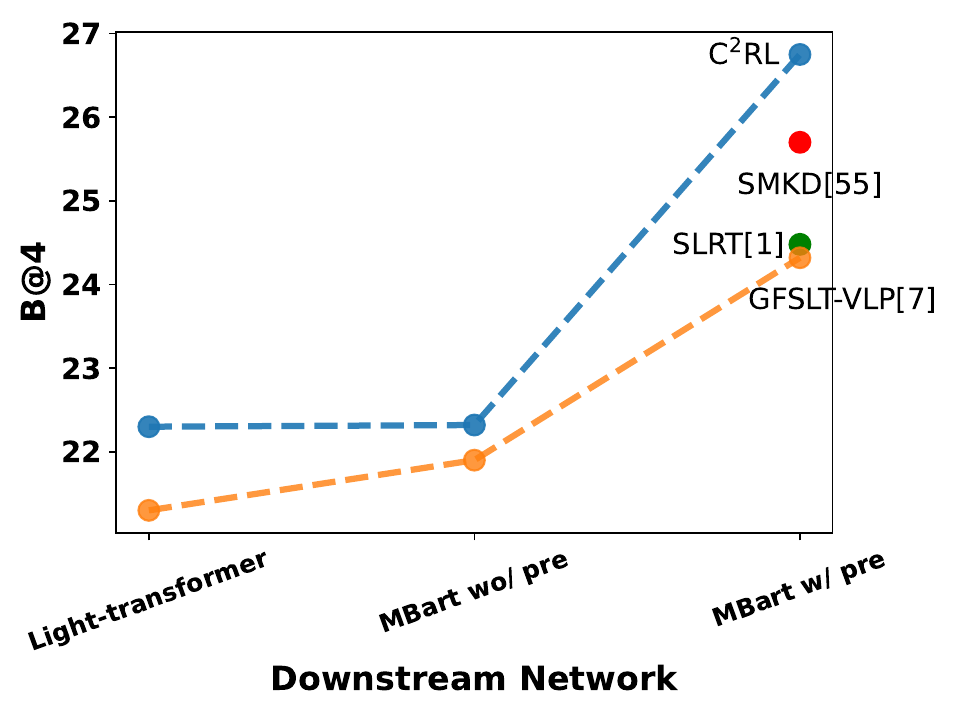}\label{subfig:slt}}
    \hfil
    \subfloat[SLRet]{\includegraphics[width=0.45\linewidth]{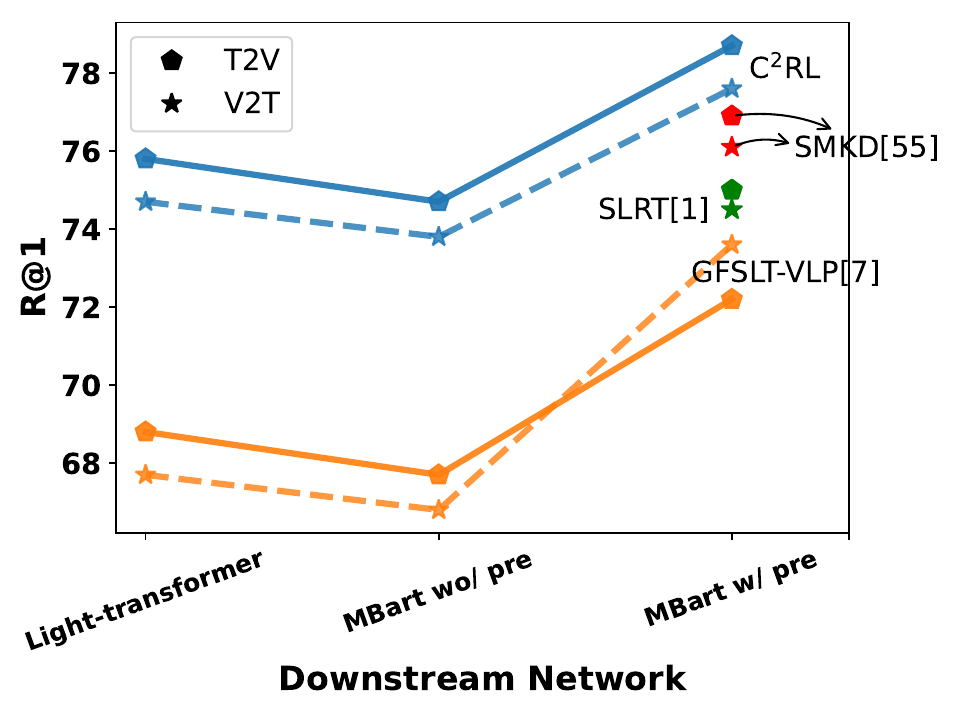} \label{subfig:slret}}
    \hfil
   \caption{The impact of the quality of pre-trained sign representations on downstream tasks. (a) Results on SLT task; (b) Results on SLRet task. \textbf{MBart wo/ pre}: random initialization.}
   \label{fig:Mbart}
\end{figure}

\paragraph{Representation Robustness} In this section, we investigate the robustness of sign language representations learned by our proposed C$^2$RL framework. We first extract sign language features using various backbones (e.g., C$^2$RL and GFSLT-VLP~\cite{Zhou_2023_ICCV}) from the training and test sets, respectively. These features are then input into downstream translation models, including a lightweight 3-layer transformer (25M) and MBart (300+M) with/without pre-training. Experimental results are detailed in Fig.\ref{fig:Mbart}. As shown, even with a lightweight 3-layer transformer as the translator, our approach maintains satisfactory performance on both downstream tasks, \ie, B@4: $>22$ and R@1: $>74$. In contrast, sign features captured by GFSLT-VLP~\cite{Zhou_2023_ICCV} demonstrate comparatively modest performance. This experiment underscores the robustness of the sign language representations acquired through C$^2$RL. Additionally, we compared the sign features extracted by two \textbf{gloss-based approaches}, SLRT~\cite{camgoz2020sign} and SMKD~\cite{hao2021self}, both of which were obtained by applying gloss for continuous sign language recognition training. It can be seen that even when compared to gloss-based sign language representations, the features learned by C$^2$RL still show competitive results. This demonstrates that C$^2$RL can perceive fine-grained concepts in sign language videos without the aid of gloss annotations, and can thus be freely extended to larger gloss-free datasets such as How2Sign and OpenASL.

\paragraph{Language Model} In the downstream tasks, we employ MBart as our pre-trained language model. In addition, here we also investigate the impact of using another language model, MT5-Base~\cite{xue-etal-2021-mt5}, which is a large language model comparable in size to MBart. As shown in Table~\ref{tab:language model}, training with MT5-Base achieves excellent performance, though it is slightly lower than MBart. This slight difference may be attributed to the strong multilingual prior knowledge in MBart. However, these results still demonstrate the strong generalization ability of the sign language features learned by C$^2$RL.


\begin{table}[!t]
    \centering
    \caption{Effect of language model.}
    \begin{tabular}{l|ccc}
    \toprule
    \multirow{2}{*}{\textbf{Model}} & {\textbf{SLT}}& {\textbf{SLRet (T2V)}}&{\textbf{SLRet (V2T)}}\\
    & B@4  &  R@1 &  R@1  \\
    \midrule
   MT5-Base~\cite{xue-etal-2021-mt5} & 25.44 & 75.7 & 74.8\\
    MBart~\cite{liu2020multilingual} & \textbf{26.75} & \textbf{78.7} & \textbf{77.6}\\
    \bottomrule
    \end{tabular}
    \label{tab:language model}
\end{table}

\section{Conclusion}
In this paper, we present an innovative approach, called C$^2$RL, to gloss-free Sign Language Representation Learning (SLRL), combining Implicit Content Learning (ICL) and Explicit Context Learning (ECL). This method offers a comprehensive understanding of sign language videos, resulting in a significant improvement in both Sign Language Translation (SLT) and Sign Language Retrieval (SLRet) tasks. While the approach is straightforward, we hope it will inspire new directions within the SLRL. Our exploration is confined to sign language video understanding due to space constraints. Nonetheless, we believe this methodology holds promise in broader areas of video understanding, such as Video Summarization and Video Captioning. We aim to further investigate these potentials in our future work.

\section*{Acknowledgments}
This work was supported by the National Key Research and Development Program of China under Grant 2021YFE0205700, Beijing Natural Science Foundation JQ23016, the External cooperation key project of Chinese Academy Sciences 173211KYSB20200002, the Science and Technology Development Fund of Macau Project 0123/2022/A3, and 0070/2020/AMJ, and CCF-Zhipu AI Large Model Project 202219.



\bibliographystyle{IEEEtran}
\bibliography{IEEEexample}

\begin{thebibliography}{10}
\providecommand{\url}[1]{#1}
\csname url@samestyle\endcsname
\providecommand{\newblock}{\relax}
\providecommand{\bibinfo}[2]{#2}
\providecommand{\BIBentrySTDinterwordspacing}{\spaceskip=0pt\relax}
\providecommand{\BIBentryALTinterwordstretchfactor}{4}
\providecommand{\BIBentryALTinterwordspacing}{\spaceskip=\fontdimen2\font plus
\BIBentryALTinterwordstretchfactor\fontdimen3\font minus \fontdimen4\font\relax}
\providecommand{\BIBforeignlanguage}[2]{{%
\expandafter\ifx\csname l@#1\endcsname\relax
\typeout{** WARNING: IEEEtran.bst: No hyphenation pattern has been}%
\typeout{** loaded for the language `#1'. Using the pattern for}%
\typeout{** the default language instead.}%
\else
\language=\csname l@#1\endcsname
\fi
#2}}
\providecommand{\BIBdecl}{\relax}
\BIBdecl

\bibitem{camgoz2020sign}
N.~C. Camgoz, O.~Koller, S.~Hadfield, and R.~Bowden, ``Sign language transformers: Joint end-to-end sign language recognition and translation,'' in \emph{Proceedings of the IEEE/CVF conference on computer vision and pattern recognition}, 2020, pp. 10\,023--10\,033.

\bibitem{Duarte_2022_CVPR}
A.~Duarte, S.~Albanie, X.~Gir\'o-i Nieto, and G.~Varol, ``Sign language video retrieval with free-form textual queries,'' in \emph{Proceedings of the IEEE/CVF Conference on Computer Vision and Pattern Recognition (CVPR)}, June 2022, pp. 14\,094--14\,104.

\bibitem{cheng2023cico}
Y.~Cheng, F.~Wei, J.~Bao, D.~Chen, and W.~Zhang, ``Cico: Domain-aware sign language retrieval via cross-lingual contrastive learning,'' in \emph{Proceedings of the IEEE/CVF Conference on Computer Vision and Pattern Recognition}, 2023, pp. 19\,016--19\,026.

\bibitem{chen2022simple}
Y.~Chen, F.~Wei, X.~Sun, Z.~Wu, and S.~Lin, ``A simple multi-modality transfer learning baseline for sign language translation,'' in \emph{Proceedings of the IEEE/CVF Conference on Computer Vision and Pattern Recognition}, 2022, pp. 5120--5130.

\bibitem{zhou2021spatial}
H.~Zhou, W.~Zhou, Y.~Zhou, and H.~Li, ``Spatial-temporal multi-cue network for sign language recognition and translation,'' \emph{IEEE Transactions on Multimedia}, vol.~24, pp. 768--779, 2021.

\bibitem{chen2022twostream}
\BIBentryALTinterwordspacing
Y.~Chen, R.~Zuo, F.~Wei, Y.~Wu, S.~LIU, and B.~Mak, ``Two-stream network for sign language recognition and translation,'' in \emph{Advances in Neural Information Processing Systems}, A.~H. Oh, A.~Agarwal, D.~Belgrave, and K.~Cho, Eds., 2022. [Online]. Available: \url{https://openreview.net/forum?id=hSxK-4KGLbI}
\BIBentrySTDinterwordspacing

\bibitem{Zhou_2023_ICCV}
B.~Zhou, Z.~Chen, A.~Clap\'es, J.~Wan, Y.~Liang, S.~Escalera, Z.~Lei, and D.~Zhang, ``Gloss-free sign language translation: Improving from visual-language pretraining,'' in \emph{Proceedings of the IEEE/CVF International Conference on Computer Vision (ICCV)}, October 2023, pp. 20\,871--20\,881.

\bibitem{yin2023gloss}
A.~Yin, T.~Zhong, L.~Tang, W.~Jin, T.~Jin, and Z.~Zhao, ``Gloss attention for gloss-free sign language translation,'' in \emph{Proceedings of the IEEE/CVF Conference on Computer Vision and Pattern Recognition}, 2023, pp. 2551--2562.

\bibitem{li2020tspnet}
D.~Li, C.~Xu, X.~Yu, K.~Zhang, B.~Swift, H.~Suominen, and H.~Li, ``Tspnet: Hierarchical feature learning via temporal semantic pyramid for sign language translation,'' \emph{Advances in Neural Information Processing Systems}, vol.~33, pp. 12\,034--12\,045, 2020.

\bibitem{zhao2021conditional}
J.~Zhao, W.~Qi, W.~Zhou, N.~Duan, M.~Zhou, and H.~Li, ``Conditional sentence generation and cross-modal reranking for sign language translation,'' \emph{IEEE Transactions on Multimedia}, vol.~24, pp. 2662--2672, 2021.

\bibitem{radford2021learning}
A.~Radford, J.~W. Kim, C.~Hallacy, A.~Ramesh, G.~Goh, S.~Agarwal, G.~Sastry, A.~Askell, P.~Mishkin, J.~Clark \emph{et~al.}, ``Learning transferable visual models from natural language supervision,'' in \emph{International Conference on Machine Learning}.\hskip 1em plus 0.5em minus 0.4em\relax PMLR, 2021, pp. 8748--8763.

\bibitem{lin2023gloss}
K.~Lin, X.~Wang, L.~Zhu, K.~Sun, B.~Zhang, and Y.~Yang, ``Gloss-free end-to-end sign language translation,'' \emph{arXiv preprint arXiv:2305.12876}, 2023.

\bibitem{duarte2021how2sign}
A.~Duarte, S.~Palaskar, L.~Ventura, D.~Ghadiyaram, K.~DeHaan, F.~Metze, J.~Torres, and X.~Giro-i Nieto, ``How2sign: a large-scale multimodal dataset for continuous american sign language,'' in \emph{Proceedings of the IEEE/CVF conference on computer vision and pattern recognition}, 2021, pp. 2735--2744.

\bibitem{shi2022open}
B.~Shi, D.~Brentari, G.~Shakhnarovich, and K.~Livescu, ``Open-domain sign language translation learned from online video,'' in \emph{EMNLP}, 2022.

\bibitem{yu2022coca}
\BIBentryALTinterwordspacing
J.~Yu, Z.~Wang, V.~Vasudevan, L.~Yeung, M.~Seyedhosseini, and Y.~Wu, ``Coca: Contrastive captioners are image-text foundation models,'' \emph{Transactions on Machine Learning Research}, 2022. [Online]. Available: \url{https://openreview.net/forum?id=Ee277P3AYC}
\BIBentrySTDinterwordspacing

\bibitem{li2022blip}
J.~Li, D.~Li, C.~Xiong, and S.~Hoi, ``Blip: Bootstrapping language-image pre-training for unified vision-language understanding and generation,'' in \emph{International Conference on Machine Learning}.\hskip 1em plus 0.5em minus 0.4em\relax PMLR, 2022, pp. 12\,888--12\,900.

\bibitem{joze2018ms}
H.~R.~V. Joze and O.~Koller, ``Ms-asl: A large-scale data set and benchmark for understanding american sign language,'' in \emph{British Machine Vision Conference}, 2019.

\bibitem{huang2018attention}
J.~Huang, W.~Zhou, H.~Li, and W.~Li, ``Attention-based 3d-cnns for large-vocabulary sign language recognition,'' \emph{IEEE Transactions on Circuits and Systems for Video Technology}, vol.~29, no.~9, pp. 2822--2832, 2018.

\bibitem{li2020transferring}
D.~Li, X.~Yu, C.~Xu, L.~Petersson, and H.~Li, ``Transferring cross-domain knowledge for video sign language recognition,'' in \emph{Proceedings of the IEEE/CVF Conference on Computer Vision and Pattern Recognition}, 2020, pp. 6205--6214.

\bibitem{koller2019weakly}
O.~Koller, N.~C. Camgoz, H.~Ney, and R.~Bowden, ``Weakly supervised learning with multi-stream cnn-lstm-hmms to discover sequential parallelism in sign language videos,'' \emph{IEEE transactions on pattern analysis and machine intelligence}, vol.~42, no.~9, pp. 2306--2320, 2019.

\bibitem{wei2020semantic}
C.~Wei, J.~Zhao, W.~Zhou, and H.~Li, ``Semantic boundary detection with reinforcement learning for continuous sign language recognition,'' \emph{IEEE Transactions on Circuits and Systems for Video Technology}, vol.~31, no.~3, pp. 1138--1149, 2020.

\bibitem{yin2023ste}
W.~Yin, Y.~Hou, Z.~Guo, and K.~Liu, ``Spatial temporal enhanced network for continuous sign language recognition,'' \emph{IEEE Transactions on Circuits and Systems for Video Technology}, 2023.

\bibitem{zuo2022c2slr}
R.~Zuo and B.~Mak, ``C2slr: Consistency-enhanced continuous sign language recognition,'' in \emph{Proceedings of the IEEE/CVF Conference on Computer Vision and Pattern Recognition}, 2022, pp. 5131--5140.

\bibitem{camgoz2018neural}
N.~C. Camgoz, S.~Hadfield, O.~Koller, H.~Ney, and R.~Bowden, ``Neural sign language translation,'' in \emph{Proceedings of the IEEE conference on computer vision and pattern recognition}, 2018, pp. 7784--7793.

\bibitem{athitsos2010large}
V.~Athitsos, C.~Neidle, S.~Sclaroff, J.~Nash, A.~Stefan, A.~Thangali, H.~Wang, and Q.~Yuan, ``Large lexicon project: American sign language video corpus and sign language indexing/retrieval algorithms,'' in \emph{sign-lang@ LREC 2010}.\hskip 1em plus 0.5em minus 0.4em\relax European Language Resources Association (ELRA), 2010, pp. 11--14.

\bibitem{zhao2024masa}
W.~Zhao, H.~Hu, W.~Zhou, Y.~Mao, M.~Wang, and H.~Li, ``Masa: Motion-aware masked autoencoder with semantic alignment for sign language recognition,'' \emph{IEEE Transactions on Circuits and Systems for Video Technology}, pp. 1--1, 2024.

\bibitem{hu2024corrnet+}
L.~Hu, W.~Feng, L.~Gao, Z.~Liu, and L.~Wan, ``Corrnet+: Sign language recognition and translation via spatial-temporal correlation,'' \emph{arXiv preprint arXiv:2404.11111}, 2024.

\bibitem{zhou2021improving}
H.~Zhou, W.~Zhou, W.~Qi, J.~Pu, and H.~Li, ``Improving sign language translation with monolingual data by sign back-translation,'' in \emph{Proceedings of the IEEE/CVF Conference on Computer Vision and Pattern Recognition}, 2021, pp. 1316--1325.

\bibitem{zhao2024conditional}
R.~Zhao, L.~Zhang, B.~Fu, C.~Hu, J.~Su, and Y.~Chen, ``Conditional variational autoencoder for sign language translation with cross-modal alignment,'' in \emph{Proceedings of the AAAI Conference on Artificial Intelligence}, vol.~38, no.~17, 2024, pp. 19\,643--19\,651.

\bibitem{LIU2024improveing}
Z.~Liu, J.~Wu, Z.~Shen, X.~Chen, Q.~Wu, Z.~Gui, L.~Senhadji, and H.~Shu, ``Improving end-to-end sign language translation with adaptive video representation enhanced transformer,'' \emph{IEEE Transactions on Circuits and Systems for Video Technology}, pp. 1--1, 2024.

\bibitem{li2023blip}
J.~Li, D.~Li, S.~Savarese, and S.~Hoi, ``Blip-2: Bootstrapping language-image pre-training with frozen image encoders and large language models,'' \emph{arXiv preprint arXiv:2301.12597}, 2023.

\bibitem{lu2019vilbert}
J.~Lu, D.~Batra, D.~Parikh, and S.~Lee, ``Vilbert: Pretraining task-agnostic visiolinguistic representations for vision-and-language tasks,'' \emph{Advances in neural information processing systems}, vol.~32, 2019.

\bibitem{chen2020uniter}
Y.-C. Chen, L.~Li, L.~Yu, A.~El~Kholy, F.~Ahmed, Z.~Gan, Y.~Cheng, and J.~Liu, ``Uniter: Universal image-text representation learning,'' in \emph{European conference on computer vision}.\hskip 1em plus 0.5em minus 0.4em\relax Springer, 2020, pp. 104--120.

\bibitem{jia2021scaling}
C.~Jia, Y.~Yang, Y.~Xia, Y.-T. Chen, Z.~Parekh, H.~Pham, Q.~Le, Y.-H. Sung, Z.~Li, and T.~Duerig, ``Scaling up visual and vision-language representation learning with noisy text supervision,'' in \emph{International conference on machine learning}.\hskip 1em plus 0.5em minus 0.4em\relax PMLR, 2021, pp. 4904--4916.

\bibitem{vlp1}
S.~Cao, G.~An, Z.~Zheng, and Z.~Wang, ``Vision-enhanced and consensus-aware transformer for image captioning,'' \emph{IEEE Transactions on Circuits and Systems for Video Technology}, vol.~32, no.~10, pp. 7005--7018, 2022.

\bibitem{vlp3}
L.~Yan, S.~Ma, Q.~Wang, Y.~Chen, X.~Zhang, A.~Savakis, and D.~Liu, ``Video captioning using global-local representation,'' \emph{IEEE Transactions on Circuits and Systems for Video Technology}, vol.~32, no.~10, pp. 6642--6656, 2022.

\bibitem{vlp5}
W.~Xu, Z.~Miao, J.~Yu, Y.~Tian, L.~Wan, and Q.~Ji, ``Bridging video and text: A two-step polishing transformer for video captioning,'' \emph{IEEE Transactions on Circuits and Systems for Video Technology}, vol.~32, no.~9, pp. 6293--6307, 2022.

\bibitem{li2019visualbert}
L.~H. Li, M.~Yatskar, D.~Yin, C.-J. Hsieh, and K.-W. Chang, ``Visualbert: A simple and performant baseline for vision and language,'' \emph{arXiv preprint arXiv:1908.03557}, 2019.

\bibitem{vlp2}
W.~Zhou and Z.~Zhou, ``Unsupervised domain adaption harnessing vision-language pre-training,'' \emph{IEEE Transactions on Circuits and Systems for Video Technology}, pp. 1--1, 2024.

\bibitem{vlp4}
J.~Feng, G.~Wang, C.~Zheng, Y.~Cai, Z.~Fu, Y.~Wang, X.-Y. Wei, and Q.~Li, ``Towards bridged vision and language: Learning cross-modal knowledge representation for relation extraction,'' \emph{IEEE Transactions on Circuits and Systems for Video Technology}, vol.~34, no.~1, pp. 561--575, 2024.

\bibitem{liu2020multilingual}
Y.~Liu, J.~Gu, N.~Goyal, X.~Li, S.~Edunov, M.~Ghazvininejad, M.~Lewis, and L.~Zettlemoyer, ``Multilingual denoising pre-training for neural machine translation,'' \emph{Transactions of the Association for Computational Linguistics}, vol.~8, pp. 726--742, 2020.

\bibitem{he2016deep}
K.~He, X.~Zhang, S.~Ren, and J.~Sun, ``Deep residual learning for image recognition,'' in \emph{Proceedings of the IEEE conference on computer vision and pattern recognition}, 2016, pp. 770--778.

\bibitem{deng2009imagenet}
J.~Deng, W.~Dong, R.~Socher, L.-J. Li, K.~Li, and L.~Fei-Fei, ``Imagenet: A large-scale hierarchical image database,'' in \emph{2009 IEEE conference on computer vision and pattern recognition}.\hskip 1em plus 0.5em minus 0.4em\relax Ieee, 2009, pp. 248--255.

\bibitem{vaswani2017attention}
A.~Vaswani, N.~Shazeer, N.~Parmar, J.~Uszkoreit, L.~Jones, A.~N. Gomez, {\L}.~Kaiser, and I.~Polosukhin, ``Attention is all you need,'' \emph{Advances in neural information processing systems}, vol.~30, 2017.

\bibitem{gutmann2010noise}
M.~Gutmann and A.~Hyv{\"a}rinen, ``Noise-contrastive estimation: A new estimation principle for unnormalized statistical models,'' in \emph{Proceedings of the thirteenth international conference on artificial intelligence and statistics}.\hskip 1em plus 0.5em minus 0.4em\relax JMLR Workshop and Conference Proceedings, 2010, pp. 297--304.

\bibitem{papineni2002bleu}
K.~Papineni, S.~Roukos, T.~Ward, and W.-J. Zhu, ``Bleu: a method for automatic evaluation of machine translation,'' in \emph{Proceedings of the 40th annual meeting of the Association for Computational Linguistics}, 2002, pp. 311--318.

\bibitem{lin2004rouge}
C.-Y. Lin, ``Rouge: A package for automatic evaluation of summaries,'' in \emph{Text summarization branches out}, 2004, pp. 74--81.

\bibitem{robbins1951stochastic}
H.~Robbins and S.~Monro, ``A stochastic approximation method,'' \emph{The annals of mathematical statistics}, pp. 400--407, 1951.

\bibitem{loshchilov2016sgdr}
I.~Loshchilov and F.~Hutter, ``Sgdr: Stochastic gradient descent with warm restarts,'' \emph{arXiv preprint arXiv:1608.03983}, 2016.

\bibitem{zhang2023sltunet}
\BIBentryALTinterwordspacing
B.~Zhang, M.~M{\"u}ller, and R.~Sennrich, ``{SLTUNET}: A simple unified model for sign language translation,'' in \emph{The Eleventh International Conference on Learning Representations}, 2023. [Online]. Available: \url{https://openreview.net/forum?id=EBS4C77p_5S}
\BIBentrySTDinterwordspacing

\bibitem{kingma2014adam}
D.~P. Kingma and J.~Ba, ``Adam: A method for stochastic optimization,'' \emph{arXiv preprint arXiv:1412.6980}, 2014.

\bibitem{uthus2023youtubeasl}
\BIBentryALTinterwordspacing
D.~Uthus, G.~Tanzer, and M.~Georg, ``Youtube-asl: A large-scale, open-domain american sign language-english parallel corpus,'' 2023. [Online]. Available: \url{https://arxiv.org/abs/2306.15162}
\BIBentrySTDinterwordspacing

\bibitem{alvarezsign}
P.~C. Alvarez, X.~G. Nieto, and L.~T. Benet, ``Sign language translation based on transformers for the how2sign dataset.''

\bibitem{tarres2023sign}
L.~Tarr{\'e}s, G.~I. G{\'a}llego, A.~Duarte, J.~Torres, and X.~Gir{\'o}-i Nieto, ``Sign language translation from instructional videos,'' in \emph{Proceedings of the IEEE/CVF Conference on Computer Vision and Pattern Recognition}, 2023, pp. 5624--5634.

\bibitem{hao2021self}
A.~Hao, Y.~Min, and X.~Chen, ``Self-mutual distillation learning for continuous sign language recognition,'' in \emph{Proceedings of the IEEE/CVF International Conference on Computer Vision}, 2021, pp. 11\,303--11\,312.

\bibitem{xue-etal-2021-mt5}
\BIBentryALTinterwordspacing
L.~Xue, N.~Constant, A.~Roberts, M.~Kale, R.~Al-Rfou, A.~Siddhant, A.~Barua, and C.~Raffel, ``m{T}5: A massively multilingual pre-trained text-to-text transformer,'' in \emph{Proceedings of the 2021 Conference of the North American Chapter of the Association for Computational Linguistics: Human Language Technologies}.\hskip 1em plus 0.5em minus 0.4em\relax Online: Association for Computational Linguistics, Jun. 2021, pp. 483--498. [Online]. Available: \url{https://aclanthology.org/2021.naacl-main.41}
\BIBentrySTDinterwordspacing

\end{thebibliography}

\vfill

\end{document}